\documentclass[journal]{IEEEtran}
\usepackage{graphicx}
\usepackage{booktabs}
\usepackage{times}
\usepackage{soul}
\usepackage{color}
\usepackage{xcolor}
\usepackage{url}
\usepackage{amsfonts,amssymb}
\usepackage[hidelinks]{hyperref}
\usepackage[utf8]{inputenc}
\usepackage[small]{caption}
\usepackage{graphicx}
\usepackage{amsmath}

\usepackage{amsthm}
\usepackage{booktabs}
\usepackage{algorithm}
\usepackage{xcolor} 
\usepackage{algorithmic}
\usepackage[switch]{lineno}
\usepackage{booktabs}
 \usepackage{multirow}
 \usepackage{pifont}
 \usepackage{bm}
 \usepackage{tabularx}
 \usepackage{adjustbox}
 \usepackage{makecell}
\usepackage{array}  
\usepackage{caption}
\newcommand{\cyq}[1]{{\color{black} #1}}

\newcommand{\ie}{{\textit{i.e.}}}

\ifCLASSINFOpdf
\else
\fi

\def\ourmethod{MVP-PCLIP}

\begin{document}
%

\title{Towards Zero-shot Point Cloud Anomaly Detection: A Multi-View Projection Framework}
%
%
%

\author{Yuqi~Cheng\IEEEauthorrefmark{1},
        Yunkang~Cao\IEEEauthorrefmark{1},
        Guoyang~Xie,~\IEEEmembership{Member, ~IEEE,},
        Zhichao~Lu,~\IEEEmembership{Member, ~IEEE,}\\
        Weiming Shen\IEEEauthorrefmark{2},~\IEEEmembership{Fellow,~IEEE,}
\thanks{Yuqi~Cheng\IEEEauthorrefmark{1} and Yunkang~Cao\IEEEauthorrefmark{1} contribute equally to this work. Weiming Shen\IEEEauthorrefmark{2} (wshen@ieee.org) is the corresponding author.}
\thanks{
Manuscript received XXXX; revised XXXX; accepted XXXX. This work was supported by Fundamental Research Funds for the Central Universities (HUST: 2021GCRC058). 
Yuqi Cheng, Yunkang Cao, and Weiming Shen are with the State Key Laboratory of Intelligent Manufacturing Equipment and Technology, Huazhong University of Science and Technology, Wuhan 430074, China. Guoyang Xie is with the Department of Intelligent Manufacturing, Contemporary Amperex Technology Limited, Ningde 352000, China and also with the Department of Computer Science, City University of Hong Kong, Hong Kong. Zhichao~Lu is with Department of Computer Science, City University of Hong Kong, Hong Kong.(e-mail: yuqicheng@hust.edu.cn; cyk\_hust@hust.edu.cn; guoyang.xie@ieee.org;luzhichaocn@gmail.com; wshen@ieee.org).

}}

\markboth{Submitted to IEEE Transactions on Systems, Man, and Cybernetics: Systems ,~Vol.~x, No.~x, February~2024}%
{Shell \MakeLowercase{\textit{et al.}}:  Bare Demo of IEEEtran.cls for IEEE Journals}
\maketitle

\IEEEpeerreviewmaketitle
\begin{abstract}

Detecting anomalies within point clouds is crucial for various industrial applications, but traditional unsupervised methods face challenges due to data acquisition costs, early-stage production constraints, and limited generalization across product categories. To overcome these challenges, we introduce the Multi-View Projection (MVP) framework, leveraging pre-trained Vision-Language Models (VLMs) to detect anomalies. Specifically, MVP projects point cloud data into multi-view depth images, thereby \cyq{translating} point cloud anomaly detection into image anomaly detection. \cyq{Following} zero-shot image anomaly detection methods, pre-trained VLMs are utilized to detect anomalies on these \cyq{depth} images. \cyq{Given that pre-trained VLMs are not inherently tailored for zero-shot point cloud anomaly detection and may lack specificity, we propose the integration of learnable visual and adaptive text prompting techniques to fine-tune these VLMs, thereby enhancing their detection performance. Extensive experiments on the MVTec 3D-AD and Real3D-AD demonstrate our proposed MVP framework's superior zero-shot anomaly detection performance and the prompting techniques' effectiveness. Real-world evaluations on automotive plastic part inspection further showcase that the proposed method can also be generalized to practical unseen scenarios.} The code is available at https://github.com/hustCYQ/MVP-PCLIP.

\end{abstract}

\section{Introduction}

\begin{figure}[t!]
\centering\includegraphics[width=\linewidth]{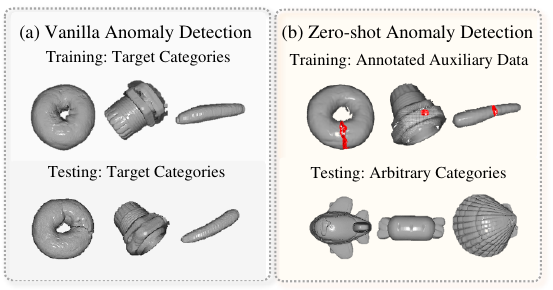}
\caption{Training schemes of different point cloud anomaly detection settings: (a) Vanilla point cloud anomaly detection and (b) Zero-shot point cloud anomaly detection. In contrast to vanilla point cloud anomaly detection, zero-shot point cloud anomaly detection leverages none samples from testing categories but annotated auxiliary data for training, aiming to build a unified model.}
\label{fig:tasksetting}
\end{figure}

\IEEEPARstart{P}{oint} clouds characterize geometric information of real-world objects~\cite{radar,CVT}, \cyq{playing a vital role in tasks such as defect detection~\cite{mvtec3d,RAD} and shape measurement~\cite{jiang2023novel,cheng2022novel,MVGR}}. Point cloud anomaly detection is emerging as a pivotal task, \cyq{with the aim to identify deviant patterns within normal point clouds that are scanned from normal products.}
Unsupervised point cloud anomaly detection~\cite{cao2024survey,cpmf} is the mainstream paradigm, training specific models for specific types of products on corresponding normal samples. However, three issues regarding industrial demands render this unsupervised scheme impractical in some cases: \cyq{\ding{182} Unsupervised anomaly detection acquires a large number of normal point cloud data for training, but collecting high precision point cloud data is very expensive}; For instance, the construction of each point cloud meeting the requirements of industrial measurement accuracy requires 1.2 days to complete with three individuals~\cite{read3d}. \cyq{\ding{183} The unsupervised learning paradigm~\cite{BiaS,PCB} is not well suited to "cold start" which means no normal point cloud data in the early stages of production. \ding{184} The unsupervised point cloud anomaly detection paradigm needs category-specific models and is not capable of generalizing to unseen categories. For the above issues, this study opts for a novel paradigm, \ie, zero-shot point cloud anomaly detection, as shown in Fig.~\ref{fig:tasksetting}, within which we aim to build a generalized model for detecting anomalies across arbitrary categories of point cloud data.}

\begin{figure}[t!]
\centering\includegraphics[width=\linewidth]{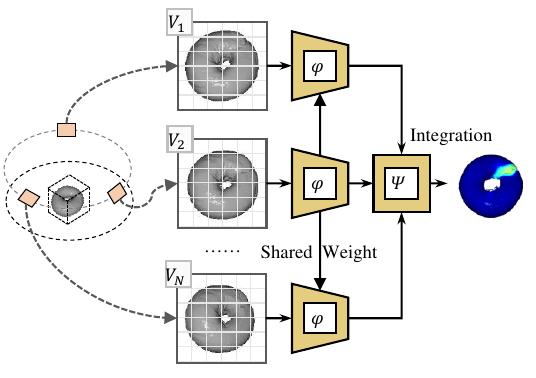}
\caption{Framework of Multi-View Projection (MVP). The origin point cloud is projected to multi-view depth images $\boldsymbol{V}_i, 1 \le i \le N$, and each depth image is delivered to a zero-shot image anomaly detection model with shared weight. The anomalies are detected by integrating the output of vision-language models.}
\label{fig:framework}
\end{figure}

\cyq{
Zero-shot anomaly detection has emerged~\cite{winclip,vand,saa, AdaCLIP} in the counterpart of point clouds, \ie, images. Most zero-shot image anomaly detection methods leverage pre-trained Vision-Language Models (VLMs) for their recognized generalization capabilities, yielding promising performance across unbounded categories. Inspired by the effectiveness of VLMs for zero-shot image anomaly detection, we propose a simple yet effective framework called Multi-View Projection (MVP) (see Fig.~\ref{fig:framework}), leveraging VLMs for point cloud anomaly detection. To meet the input conditions of the VLMs, MVP projects the testing point cloud into multi-view depth images and then applies VLMs on these projected images. 
Given that multi-view depth images offer a comprehensive representation of the point cloud, the detection outcomes derived from these images can disclose anomalies within the point cloud.
}

\begin{figure}[h!]
\centering\includegraphics[scale=1.0]{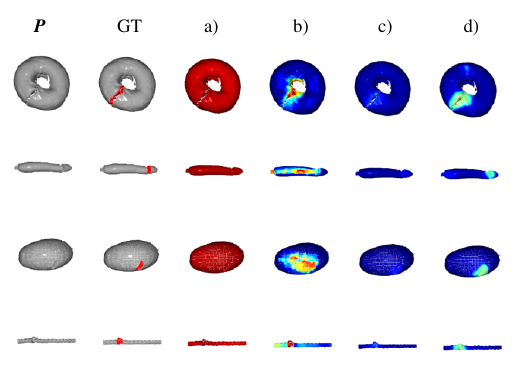}
\caption{{Comparisons between simple integration of the proposed MVP and existing zero-shot image anomaly detection methods and the proposed method MVP-PCLIP.} (a) MVP-WinCLIP, (b) MVP-SAA, (c) MVP-APRIL-GAN, (d) MVP-PCLIP. 
}
\label{fig:simvis}
\end{figure}

With the proposed MVP framework, existing zero-shot image anomaly detection methods can be seamlessly adapted for point clouds. However, as depicted in Fig.~\ref{fig:simvis}, the integration of MVP with existing zero-shot image anomaly detection methods, namely MVP-WinCLIP~\cite{winclip}, MVP-SAA~\cite{saa}, and MVP-APRIL-GAN~\cite{vand}, struggles to detect point cloud anomalies accurately. We attribute the subpar performance on point clouds to the inherent domain gap from two aspects. 
Firstly, a domain gap exists between the original pretraining and anomaly detection tasks. Existing zero-shot image anomaly detection methods tackle this gap through text prompts~\cite{winclip,saa}, but these hand-crafted text prompts are found to be ineffective for point clouds. Consequently, we further introduce learnable text prompts to bridge this gap. On the other hand, VLMs are predominantly trained on RGB images, introducing another domain gap to the rendered depth images. We develop learnable visual prompts to address this gap. After applying the learnable text and visual prompts to a widely employed VLM, CLIP, we present an improved method named MVP-PCLIP. To optimize these prompts automatically, we follow an existing supervised zero-shot anomaly detection paradigm~\cite{vand}. As evidenced in Fig.~\ref{fig:simvis}, MVP-PCLIP significantly outperforms other zero-shot anomaly detection alternatives through the introduced learnable prompts, demonstrating that the existing domain gaps are better mitigated.

In summary, the contributions of this study are summarized as follows:

\begin{itemize}
    \item To the best of our knowledge, we are the first to introduce zero-shot point cloud anomaly detection, aiming to construct a generalized anomaly detection model for point clouds from arbitrary categories. 
    \item We propose a Multi-View Projection framework for zero-shot point cloud anomaly detection, which could seamlessly adapt existing zero-shot image anomaly detection methods for zero-shot point cloud anomaly detection.
    \item We identify the domain gap problem by simply adapting zero-shot image anomaly detection for zero-shot point cloud anomaly detection. To bridge this gap, we propose an improved method \ourmethod{}, introducing two types of prompts specialized for CLIP's text and visual branches. \ourmethod{} significantly improves the suitability of CLIP for zero-shot point cloud anomaly detection and outperforms other methods.
    \item We establish a comprehensive benchmark for zero-shot point cloud anomaly detection, which greatly facilitates the development of this field. 
\end{itemize}

\cyq{The structure of this paper is as follows. Section~\ref{sec:related-work} provides a review of the related work. In Section~\ref{sec:definition}, the research scheme is outlined. Section~\ref{sec:mvp} presents the proposed MVP framework, followed by a detailed explanation of the enhanced method, \ourmethod{}, in Section~\ref{sec:mvp-pclip}. The experimental evaluation of the proposed approach is discussed in Section~\ref{sec:exp}. Finally, the conclusion is drawn in Section~\ref{sec:conclusion}.}

\section{Related Work}\label{sec:related-work}
In this section, we review topics closely related to the technical aspects of this paper, including image anomaly detection, point cloud anomaly detection, and prompt learning.

\subsection{Image Anomaly Detection}
Unsupervised anomaly detection is the mainstream paradigm in industrial image anomaly detection field. Specifically, unsupervised anomaly detection algorithms can be divided into five categories, including methods based on reconstruction~\cite{Huang_VAD,jiang2022masked,matsubara2020deep,chen2023easynet}, one-class classification~\cite{massoli2021mocca,xie2023iad,liu2024deep}, teacher-student structure~\cite{cdo,RD,RD++}, normalizing flow~\cite{rudolph2021same} and memory bank~\cite{patchcore}. However, the expensive acquisition cost of normal data and the ``cold start'' problem result in limited or even no available samples in anomaly detection~\cite{xie2023pushing,winclip}, which cannot meet the requirements of large dataset quantities for training in the unsupervised scheme.

Recently, zero-shot anomaly detection~\cite{winclip} methods have shown their potential in manufacturing. WinCLIP~\cite{winclip} is the first one to employ CLIP, a widely exploited VLM, for zero-shot image anomaly detection. WinCLIP~\cite{winclip} designs some hand-crafted text prompts to harness the generalization capability of CLIP. To adapt the zero-shot classification principle of CLIP into anomaly segmentation, WinCLIP proposes a window-based strategy for fine-grained segmentation. Following WinCLIP, APRIL-GAN~\cite{vand} utilizes the shallow and deep features of the CLIP image encoder to improve the accuracy of anomaly detection. Then, SAA~\cite{saa} collaboratively assembles different VLMs for anomaly detection and improves their performance in anomaly detection via hybrid prompt regularization. With the incorporation of VLMs, existing zero-shot image anomaly detection methods have achieved promising anomaly detection performance on novel categories. Inspired by the success of zero-shot image anomaly detection methods, we aim to apply zero-shot anomaly detection techniques to point clouds. To achieve this, we propose a novel MVP framework that leverages VLMs by rendering point clouds into multi-view images.

\subsection{Point Cloud Anomaly Detection}

\cyq{Unsupervised learning methods also play a vital role in point cloud anomaly detection. Similar to image anomaly detection, variational autoencoder~\cite{3dvae} and teacher-student structure models~\cite{3dst,3dst_self_supervised,ast} are utilized to calculate the anomaly score of point clouds. 

More recently, CPMF~\cite{cpmf} proposes complementary pseudo multimodal features that incorporate the features generated by handcrafted point cloud descriptions and pre-trained 2D neural networks, achieving outstanding performance.}

\cyq{The challenges posed by the ``cold start'' problem and the significantly higher acquisition cost of normal data for point clouds, compared to images, underscore the necessity of exploring zero-shot point cloud anomaly detection methods. The rapid advancement in zero-shot image anomaly detection is largely attributed to the utilization of off-the-shelf VLMs, which demonstrate remarkable generalization capabilities. In this context, we aim to leverage these VLMs for point cloud data. To this end, we introduce the MVP framework, which transforms testing point clouds into multi-view depth images, enabling the adaptation of existing zero-shot image anomaly detection techniques to point clouds. Considering the domain gaps between current VLMs and zero-shot point cloud anomaly detection, we further propose MVP-PCLIP, which incorporates learnable prompts into the employed VLM (CLIP), thereby achieving superior performance in zero-shot point cloud anomaly detection.}

\subsection{Prompt Learning}
VLMs have demonstrated remarkable success in many fields. CLIP~\cite{clip} is a pioneering VLM, having demonstrated exceptional capabilities in zero-shot \cyq{tasks}, such as image segmentation~\cite{xu2022groupvit}, object detection~\cite{li2022grounded}. Although VLMs such as CLIP have exhibited strong generalization capacities, incorporating visual and text prompts can further refine their performance in specific tasks. AdaptFormer~\cite{chen2022adaptformer} integrates lightweight modules as prompts into vision-language models to improve the efficiency of action recognition benchmarks. From the perspective of model fine-tuning, VPT~\cite{jia2022visual} and ViPT~\cite{zhu2023visual} freeze the network weights and integrate a small number of learnable parameters as visual prompts to fine-tune the network, thereby improving the VLMs' downstream task performance. Considering the crucial role of text description in visual-language models, manually designed text prompts for specific objects are proposed to improve downstream task performance~\cite{winclip}. More recently, the learnable prompt has been proposed like CoOp~\cite{CoOp} that introduces the automatic prompts representing the specific prompt for the downstream task as a trainable continuous vector. Moreover, MaPLe~\cite{maple} extends CoOp to multi-modal prompts that enhance the information transmission between text and images, allowing the model to learn more contextual and semantic information during training. 

\cyq{
In the context of zero-shot point cloud anomaly detection, the domain gap between the training data of VLMs and the target anomaly data adversely affects detection performance. Additionally, the discrepancy between RGB images and rendered multi-view depth images further amplifies this negative impact. To mitigate these challenges, this paper introduces MVP-PCLIP, which employs visual and textual prompts to extract more appropriate image feature representations and generate accurate textual descriptions.
}

\section{Framework}\label{sec:framework}

\subsection{Problem Definition}\label{sec:definition}

Point cloud anomaly detection aims to compute the object-wise anomaly score $\xi \in [0,1]$ and point-wise anomaly map $\boldsymbol{A} \in \mathbf{R}^{n \times 1}$ for any point cloud $\boldsymbol{P} \in \mathbf{R}^{n\times3}$ with $n$ points. Higher values for $\xi$ and $\boldsymbol{A}$ indicate increased anomaly levels.

This paper addresses the zero-shot point cloud anomaly detection task, wherein point cloud anomaly detection models $f_\theta$ are trained to detect anomalies for unseen categories. Zero-shot point cloud anomaly detection offers versatility across various categories but presents challenges as no prior knowledge about the targeted categories is available. We adopt a supervised training paradigm for this task, aiming to build a general zero-shot point cloud anomaly detection model, leveraging common knowledge inherent in extensive existing point cloud anomaly detection datasets. Formally, an annotated auxiliary point cloud anomaly detection dataset $\mathcal{X}_{train}$, along with its object-wise labels $\{ \xi \}_{train}$ and point-wise masks $\{ \boldsymbol{A} \}_{train}$, is provided for training $f_\theta$. Subsequently, $f_\theta$ is exploited to generate the object-wise anomaly score $\xi$ and point-wise anomaly map $\boldsymbol{A}$ for point clouds from unseen categories, thereby achieving zero-shot point cloud anomaly detection.

\subsection{Multi-View Projection}\label{sec:mvp}

\cyq{To achieve zero-shot point cloud anomaly detection, this paper proposes the MVP framework (shown in Fig.~\ref{fig:framework}) that connects point clouds with pre-trained VLMs}. Within the proposed framework, original point clouds are projected into multi-view depth images.  Subsequently, VLMs are employed to perform zero-shot image anomaly detection for individual depth images. Finally, the anomaly scores of individual projected depth images are integrated for the final results. The detailed expressions are as follows.

\subsubsection{Point Cloud Projection}

\begin{figure}[t!]
\centering\includegraphics[scale=1.0]{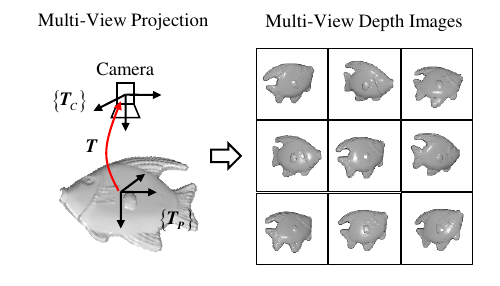}
\caption{The projection of point cloud from multiple views.}
\label{fig:projection}
\end{figure}

The point cloud projection is illustrated in Fig.~\ref{fig:projection}. $\boldsymbol{T}$ is the homogeneous transformation matrix from the coordinate system of the point cloud $\{ \boldsymbol{T}_p \}$ to the camera coordinate system $\{ \boldsymbol{T}_c \}$. Let $\boldsymbol{P}^i$ be the $i$-th point in $\boldsymbol{P}$, and $\boldsymbol{V}^i$ be the corresponding point in the pixel coordinate system. The projection relationship between $\boldsymbol{P}^i$ and $\boldsymbol{V}^i$ is given by:

\begin{equation}
[\boldsymbol{V}^i,1]^{\rm T}=\frac{1}{z_i}\boldsymbol{KT}[\boldsymbol{P}^i,1]^{\rm T}\label{F1}
\end{equation}

\noindent where $z_i$ is the Z-coordinate of $\boldsymbol{P}^i$ in $\{ \boldsymbol{T}_c \}$, and $\boldsymbol{K}$ is the intrinsic matrix of the camera.

Setting different poses $\boldsymbol{T}_k,1\le k \le N_v$, multiple depth images $\boldsymbol{V}_k,1\le k \le N$ can be generated after projection and rendering, as shown in Fig.~\ref{fig:projection}. Let $\boldsymbol{V}_k^i$ be the corresponding point of $\boldsymbol{P}^i$ in $\boldsymbol{V}_k$, and the projection relationship between $\boldsymbol{P}^i$ and $\boldsymbol{V}_k^i$ is given by:

\begin{equation}
[\boldsymbol{V}^i_k,1]^{\rm T}=\frac{1}{z_i}\boldsymbol{KT}_k[\boldsymbol{P}^i,1]^{\rm T}\label{F2}
\end{equation}

Define $\mathrm{Pos_k}: \boldsymbol{V}_k \longmapsto \boldsymbol{P}$ as the mapping from image pixels to points in the point cloud:

\begin{equation}
\boldsymbol{P}=\mathrm{Pos_k}(\boldsymbol{V}_k)=\boldsymbol{R}_k^{-1}(z_i \boldsymbol{K}^{-1} \boldsymbol{V}^i_{k}- \boldsymbol{t}_k)\label{F2}
\end{equation}

\noindent where $\boldsymbol{R}_k$ is the rotation matrix of $\boldsymbol{T}_k$, and $\boldsymbol{t}_k$ is the translation vector of $\boldsymbol{T}_k$. $\mathrm{Pos_k}$ function aligns the features of points in the subsequent integration process.

\subsubsection{Single-View Anomaly Detection}

After multi-view projection, MVP employs a shared zero-shot image anomaly detection model $\varphi$ across all views to perform single-view anomaly detection. For any depth image $V_i$, the output of $\varphi$ can be obtained as follows:

\begin{equation}
\boldsymbol{C}_i = \varphi(V_i)\label{F3}
\end{equation}

\noindent where $\boldsymbol{C}_i$ denotes to the detection results of $i$-th depth images. 

\begin{figure*}[t!]
\centering\includegraphics[width=\linewidth]{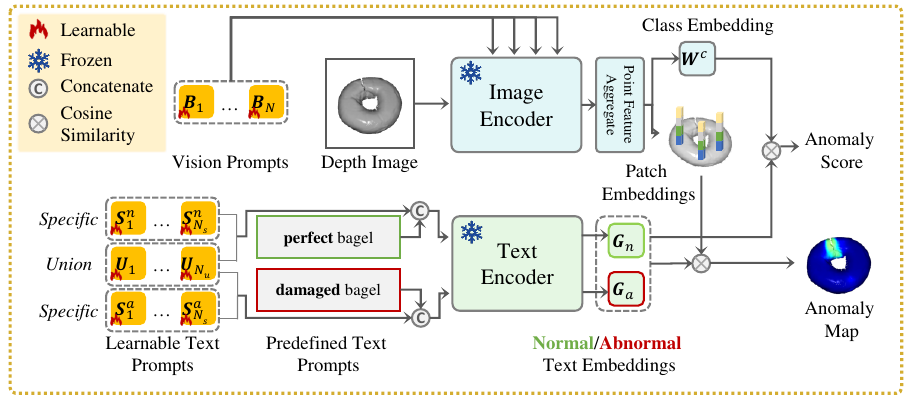}
\caption{The framework of the proposed MVP-PCLIP. The depth images projected from multiple perspectives are input into image encoder with visual prompt to extract image features. The text prompts are constructed by combining the learnable text prompts and predefined text prompts, and text features are extracted through the text encoder. Next, the point features are aggregated through the projection relationship between the point cloud and images. Finally, the similarity between points and text features is leveraged for object-wise and point-wise anomaly detection.}
\label{fig:method_overview}
\end{figure*}

\subsubsection{Integration}

Multiple outputs $\boldsymbol{C}_i , 1\le i\le N$ are obtained for individual depth images through single-view anomaly detection. Let $\Psi$ be the integration function to fuse these outcomes for final point cloud anomaly detection,

\begin{equation}
\xi,\boldsymbol{A} = \Psi (\{ \boldsymbol{C}_1, \boldsymbol{C}_2, \cdots, \boldsymbol{C}_N \})\label{F3}
\end{equation}

\noindent the anomaly score $\xi$ and anomaly map $\boldsymbol{A}$ are obtained.
 
\section{MVP-PCLIP}\label{sec:mvp-pclip}
\subsection{Overview}
\cyq{

As previously discussed, while the MVP framework facilitates the application of zero-shot image anomaly detection methods to zero-shot point cloud anomaly detection, its performance remains suboptimal due to the domain gap between the general knowledge of VLMs and the specific anomaly understanding required for the target data. To address this challenge, this paper introduces MVP-PCLIP, which prompts and integrates CLIP into our MVP framework, as shown in Fig.~\ref{fig:method_overview}. Specifically, following previous zero-shot methods~\cite{vand,AdaCLIP}, MVP-PCLIP leverages CLIP to compute the similarities between image and text prompts regarding normal and abnormal states for anomaly detection. During the detection process, MVP-PCLIP first extracts image features from projected depth images and text features from the designed prompts using the CLIP image and text encoders. Object-wise and point-wise anomaly scores are then computed by measuring the similarity between the image and text features, and larger similarities to the abnormal text features indicate a higher abnormal level. 
To mitigate the domain gap, a Key Layer Visual Prompt (KLVP) is incorporated into the image encoder to adjust the output feature distributions at key layers. Additionally, an Adaptive Text Prompt (ATP) is introduced to refine text descriptions, enhancing the identification of abnormal points. ATP dynamically optimizes the text description by integrating both specific and union text prompts.
}

\subsection{Key Layer Visual Prompt}
The features obtained in the pre-training image encoder are often strongly correlated with the distribution of the pre-training dataset, \cyq{resulting in a domain gap to the anomaly detection task.} Therefore, we consider incorporating learnable parameters for \cyq{fine-tuning the pre-training image encoder to bridge the domain gap.}

\cyq{The features extracted by different layers of the image encoder represent varying semantic information.} To fuse multi-scale features of the image, key layers are selected, denoted as  $L_{k_i},~k_i\in \{k_1,k_2,\cdots,k_m\},1 \le k_i \le N_l$, where $N_l$ is the total layer number of the image encoder. \cyq{The outputs of key layers are used for anomaly detection}. Define $\boldsymbol{W}_i$ as the patch embedding obtained by \textit{i}-th transformer layer:
\begin{equation}
\boldsymbol{W}_i = L_i(\boldsymbol{W}_{i-1})\label{F9}
\end{equation} 
When the learnable parameters are added at the transformer layers, the output is updated as:

\begin{equation}
\left\{
\begin{aligned}
\boldsymbol{W}_j &= L_j(\boldsymbol{W}_{j-1}), \quad & \text{if}\ j\notin \{k_1,k_2,\cdots,k_m\}  \\
\tilde{\boldsymbol{W}}_j &= L_j(\boldsymbol{W}_{j-1},\boldsymbol{B}_j), \quad & \text{if}\ j \in \{k_1,k_2,\cdots,k_m\}\\
\end{aligned}
\right.\label{F10}
\end{equation}
where $\boldsymbol{B}_j$ is the learnable parameters applied in the \textit{j}-th transformer layer. $\tilde{\boldsymbol{W}}_j$ is the feature that is more conducive to describing target image distribution.

\cyq{By introducing visual prompts}, we hypothesize that the image encoder focuses more on defect regions, thereby enhancing the zero-shot generalization capabilities.

\subsection{Adaptive Text Prompt}

\cyq{Two text prompts are crafted to describe normal or abnormal images/points and the anomaly score can be calculated by assessing the similarity between the features of images/points and text prompts}. Let \textit{State} $\in$ \{\textit{perfect}, \textit{normal}, \textit{damaged}, \textit{abnormal}, etc.\} denote the state description, and \textit{CLS} $\in$ \{\textit{bagel}, \textit{carrot}, \textit{airplane}, etc.\} represents the object category. The naive text prompt is constructed as follows:
\begin{equation}
T_{\text{prompt}} = h([\text{State}, \text{CLS}])\label{F11}
\end{equation} 
where $h([\text{State}, \text{CLS}])$ forms a complete text prompt, such as ``a photo of a perfect bagel'' for normal description $t_n$ and "a photo of a damaged bagel" for abnormal description $t_a$.

Indeed, the performance of MVP-PCLIP depends on the text prompting. However, the text prompting tailored for point cloud anomaly detection is challenging to design. Considering a text prompt can be not only a discrete \cyq{explicit human language expression but also a continuous abstract vector.} \cyq{Therefore, we consider using learnable parameters to dynamically generate text prompts:} 
\begin{equation}
\tilde{T}_{prompt} = \tilde{h}([State,CLS])\label{F11}
\end{equation} 
where $\tilde{h}(x) = cat(\boldsymbol{U},\boldsymbol{S},State,CLS)$, \cyq{$\boldsymbol{U}$ is the learnable parameters as union text prompts to describe the prefix for images such as "a photo of", $\boldsymbol{S}$ is the learnable parameters as specific text prompt to describe the defect state such as "abnormal" and "good". The $\boldsymbol{U}$ is both shared in two text prompts but $\boldsymbol{S}$ is not. }
For example, normal description $t_n$ and abnormal description $t_a$ for "bagel" can be changed to:
\begin{equation}
t_n = [U_1]\cdots[U_{n_u}][S_1^n]\cdots[S_{n_s}^n]~\text{perfect bagel}\label{F11}
\end{equation} 
\begin{equation}
t_a = [U_1]\cdots[U_{n_u}][S_1^a]\cdots[S_{n_s}^a]~\text{damaged bagel}\label{F11}
\end{equation} 
where $n_u$ is the number of learnable parameters in \cyq{the union text prompts} and $n_s$ is the number of learnable parameters in \cyq{the specific text prompts}. $S_i^n$ and $S_i^a$ are different parameters used for $t_n$ and $t_a$, respectively.

\subsection{Point Cloud Anomaly Detection}
$\boldsymbol{\tilde{W}}_{jk}$ is the $k$-th transformer layer output of the image encoder $f$ in $j$-th depth image. Expand the channel of $\boldsymbol{\tilde{W}}_{jk}$ into the feature map $\boldsymbol{F}_{jk}$. The next step is to aggregate the features of each point from the multi-view depth images. The point features $\boldsymbol{P}^i_{jk}$ of point $\boldsymbol{P}^i$ on the $j$-th depth image can be obtained:
\begin{equation}
\boldsymbol{P}_{jk}^i = Pos_j(\boldsymbol{F}_{jk}^i) \label{F11}
\end{equation} 
Define $\boldsymbol{\tilde{P}}$ as the point features aggregated from multi-view depth images, and $i$-th point features $\boldsymbol{\tilde{P}^i}$ can be obtained:
\begin{equation}
\boldsymbol{\tilde{P}^i} = \frac{1}{mN}\sum_j^N \sum_l^{m} \boldsymbol{P}_{jk_l}^i \label{F11}
\end{equation} 

The text features $\boldsymbol{G}$ are acquired by text encoder $g$:
\begin{equation}
\boldsymbol{G}=g(t_n,t_a) \label{F11}
\end{equation} 

The anomaly score $\xi$ is calculated using the class token $\tilde{\boldsymbol{W}}_i^{clc}$ extracted from the $i$-th image and text features $\boldsymbol{G}$ : 
\begin{equation}
\xi = softmax((\frac{1}{N}\sum_i^N \tilde{\boldsymbol{W}}_i^{clc})\boldsymbol{G}^{\rm T})\label{F12}
\end{equation}

The point-wise anomaly map $\boldsymbol{A}$ can be calculated by:
\begin{equation}
\boldsymbol{A} = softmax(\tilde{\boldsymbol{P}} \boldsymbol{G}^{\rm T})\label{F13}
\end{equation}

During the training process, the loss function $\mathcal{L}$ is:
\begin{multline}
    \mathcal{L} = IoU(\boldsymbol{A},\boldsymbol{A}_{gt})+Focal(\boldsymbol{A},\boldsymbol{A}_{gt})
\\+CrossEntroy(\xi,\xi_{gt})\label{F13}
\end{multline}
\cyq{where $\boldsymbol{A}_{gt}$ and $\xi_{gt}$ are the ground true anomaly map and anomaly score, respectively. IoU loss and Focal loss are used for point-wise anomaly detection task and CrossEntropy loss is chosen for object-wise anomaly detection task.}

\section{Experiment}\label{sec:exp}
\subsection{Experimental Settings}
\subsubsection{Dataset}

The experiments are conducted on the MVTec 3D-AD dataset~\cite{mvtec3d} and Real3D-AD dataset~\cite{read3d}. MVTec 3D-AD and Real3D-AD contain high-resolution scans of 10 and 12 different object categories obtained from industrial 3D sensors, respectively. 
\cyq{For our zero-shot point cloud anomaly detection task, we utilize only the test sets from the original datasets, which consist of point clouds annotated at the point level. Background points of MVTec 3D-AD dataset are filtered out, and we convert the Real3D-AD dataset to a form similar to the MVTec 3D-AD dataset. Subsequently, both datasets are projected into multi-view depth images, utilizing nine depth images in total.
 }

\subsubsection{Implementation Details}

The image encoder and text encoder originate from CLIP \footnote{https://github.com/openai/CLIP}, with a 224-pixel input image size and a 768-dimensional embedding feature. The 6-th, 12-th, 18-th and 24-th layers of the image encoder are chosen as key layers. The number of union/specific text prompts is set to 8/4.  
\cyq{For zero-shot anomaly detection evaluations, the proposed model undergoes training on MVTec-3D and testing on Real-3D, and vice versa}. The training epoch is set to 3, with a learning rate of 0.0005. \cyq{The model inference is conducted on the GPU RTX A6000, employing PyTorch version 1.7.}

\subsubsection{Evaluation Metrics}

The evaluation of anomaly detection performance includes Area Under the Receiver Operating Characteristic curve (AUROC), Maximal F1 score (Max-F1), and Average Precision (AP). We denote them as Object-AUROC (O-R), Object-Max-F1 (O-F) and Object-AP (O-P) in object-wise assessment and Point-AUROC (P-R), Point-Max-F1 (P-F) and Point-AP (P-P) in point-wise assessment. It is imperative to note that the point-level evaluation exclusively concentrates on foreground points. This intentional focus aims to prevent potential misinterpretation, as evaluating all points might result in artificially inflated metrics.

\subsection{Comparison Studies}
\subsubsection{Baseline Methods Selection}

\cyq{To thoroughly evaluate the efficacy of zero-shot point cloud anomaly detection, this paper systematically assesses various methods, including CPMF~\cite{cpmf}, PointMAE~\cite{pointmae}, MVP-WinCLIP, MVP-SAA, and MVP-APRIL-GAN. CPMF, an unsupervised method currently achieving the best anomaly detection performance for point clouds, is chosen as the benchmark for comparison with zero-shot approaches. PointMAE, a feature extractor specifically designed for point clouds, is integrated into the zero-shot scheme in this study. MVP-WinCLIP, MVP-SAA, and MVP-APRIL-GAN are developed by incorporating well-established zero-shot image anomaly detection methods, namely WinCLIP~\cite{winclip}, SAA~\cite{saa}, and APRIL-GAN~\cite{vand}, into the proposed MVP framework.
}

\subsubsection{Point-wise Anomaly Detection}
Quantitative results for the MVTec 3D-AD dataset and Real3D-AD dataset are detailed in Table~\ref{table:pointmvtec} and Table~\ref{table:pointreal}. MVP-PCLIP and MVP-APRIL-GAN adopt the supervised learning scheme and have also achieved competitive results in both datasets, which proves the effectiveness of \cyq{the} supervised learning scheme in auxiliary datasets. Weakening the gap using proposed visual/text prompts, MVP-PCLIP shows the best performance in most categories. In comparison with CPMF, our method demonstrates competitive performance metrics, achieving P-R at 87.5
\%, P-F at 33.6\%, and P-P at 26.9\% in MVTec 3D-AD dataset and achieves P-R at 84.6\%, P-F at 23.8\%, and P-P at 17.9\% in Real3D-AD dataset. Notably, MVP-PCLIP, MVP-APRIL-GAN, MVP-SAA, and MVP-WinCLIP all surpass unsupervised methods CPMF in all or part of the indicators, underscoring the effectiveness and progressiveness of MVP framework. This indicates the framework's capacity to leverage powerful VLMs for images to point cloud anomaly detection.

\begin{table*}[]
\setlength{\tabcolsep}{8pt}
\centering
\caption{Quantitative Point-wise Results on MVTec 3D-ad Dataset. The Three Indicators In Parentheses Are P-R, P-F And P-P, Respectively. The Best Is In \textbf{Bold}, And The Second Best In \underline{Underlined}. (\%)}
\setlength\tabcolsep{5.0pt}
\resizebox{\linewidth}{!}{
\begin{tabular}{@{}ccccccc@{}}
\toprule[1.5pt]

Category   
& \begin{tabular}[c]{@{}c@{}}CPMF (Unsupervised)~\cite{cpmf}\\ \textit{PR'2024}\end{tabular}
& \begin{tabular}[c]{@{}c@{}}PointMAE~\cite{pointmae}\\ \textit{ECCV'2022}\end{tabular}
& \begin{tabular}[c]{@{}c@{}}MVP-WinCLIP~\cite{winclip}\\ \textit{CVPR'2023}\end{tabular}
& \begin{tabular}[c]{@{}c@{}}MVP-SAA~\cite{saa}\\ \textit{CVPRW'2023}\end{tabular}
& \begin{tabular}[c]{@{}c@{}}MVP-APRIL-GAN~\cite{vand}\\ \textit{CVPRW'2023}\end{tabular}
& \ourmethod{} \\ \midrule

Bagel        
& (96.6\begin{tiny}±0.0\end{tiny}, 53.0\begin{tiny}±0.0\end{tiny}, 55.2\begin{tiny}±0.0\end{tiny}) 
& (50.0\begin{tiny}±0.0\end{tiny}, 2.6\begin{tiny}±0.0\end{tiny}, 1.3\begin{tiny}±0.0\end{tiny})  
& (58.1\begin{tiny}±0.0\end{tiny}, 3.4\begin{tiny}±0.0\end{tiny}, 1.7\begin{tiny}±0.0\end{tiny})   
& (90.1\begin{tiny}±0.0\end{tiny}, 32.3\begin{tiny}±0.0\end{tiny}, 25.5\begin{tiny}±0.0\end{tiny}) 
& (\underline{95.9}\begin{tiny}±0.1\end{tiny}, \underline{38.8}\begin{tiny}±0.8\end{tiny}, \underline{31.6}\begin{tiny}±1.2\end{tiny}) 
& (\textbf{97.1}\begin{tiny}±0.1\end{tiny}, \textbf{41.7}\begin{tiny}±0.6\end{tiny}, \textbf{35.0}\begin{tiny}±0.8\end{tiny})                                                                    \\
Cable\_Gland 
& (92.8\begin{tiny}±0.0\end{tiny}, 34.5\begin{tiny}±0.0\end{tiny}, 30.6\begin{tiny}±0.0\end{tiny})      
& (50.0\begin{tiny}±0.0\end{tiny}, 3.1\begin{tiny}±0.0\end{tiny}, 1.6\begin{tiny}±0.0\end{tiny})  
& (61.4\begin{tiny}±0.0\end{tiny}, 5.8\begin{tiny}±0.0\end{tiny}, 2.7\begin{tiny}±0.0\end{tiny})   
& (56.9\begin{tiny}±0.0\end{tiny}, 4.1\begin{tiny}±0.0\end{tiny}, 1.8\begin{tiny}±0.0\end{tiny})   
& (\underline{76.8}\begin{tiny}±1.8\end{tiny}, \underline{15.0}\begin{tiny}±1.5\end{tiny}, \underline{7.4}\begin{tiny}±1.6\end{tiny})  
& (\textbf{82.2}\begin{tiny}±0.5\end{tiny}, \textbf{20.4}\begin{tiny}±0.8\end{tiny}, \textbf{12.1}\begin{tiny}±0.7\end{tiny})                                                                    \\
Carrot       
& (95.5\begin{tiny}±0.0\end{tiny}, 49.2\begin{tiny}±0.0\end{tiny}, 48.3\begin{tiny}±0.0\end{tiny})      
& (50.2\begin{tiny}±0.0\end{tiny}, 5.7\begin{tiny}±0.0\end{tiny}, 3.0\begin{tiny}±0.0\end{tiny})  
& (73.5\begin{tiny}±0.0\end{tiny}, 12.4\begin{tiny}±0.0\end{tiny}, 8.7\begin{tiny}±0.0\end{tiny})  
& (88.7\begin{tiny}±0.0\end{tiny}, 36.5\begin{tiny}±0.0\end{tiny}, 29.4\begin{tiny}±0.0\end{tiny})
& (\underline{91.9}\begin{tiny}±0.3\end{tiny}, \underline{36.6}\begin{tiny}±0.3\end{tiny}, \underline{29.9}\begin{tiny}±0.2\end{tiny})
& (\textbf{94.4}\begin{tiny}±0.2\end{tiny}, \textbf{40.7}\begin{tiny}±0.5\end{tiny}, \textbf{34.2}\begin{tiny}±0.7\end{tiny})                                                                    \\
Cookie       
& (92.8\begin{tiny}±0.0\end{tiny}, 51.9\begin{tiny}±0.0\end{tiny}, 47.0\begin{tiny}±0.0\end{tiny})    
& (50.0\begin{tiny}±0.0\end{tiny}, 4.6\begin{tiny}±0.0\end{tiny}, 2.4\begin{tiny}±0.0\end{tiny}) 
& (71.8\begin{tiny}±0.0\end{tiny}, 11.2\begin{tiny}±0.0\end{tiny}, 6.0\begin{tiny}±0.0\end{tiny})
& (76.0\begin{tiny}±0.0\end{tiny}, \textbf{39.7}\begin{tiny}±0.0\end{tiny}, \textbf{32.2}\begin{tiny}±0.0\end{tiny}) 
& (\textbf{85.4}\begin{tiny}±0.3\end{tiny}, \underline{33.0}\begin{tiny}±1.2\end{tiny}, \underline{27.1}\begin{tiny}±2.1\end{tiny}) 
& (\underline{84.7}\begin{tiny}±0.3\end{tiny}, 31.8\begin{tiny}±1.4\end{tiny}, 25.2\begin{tiny}±1.4\end{tiny})                                                                    \\
Dowel     
& (89.0\begin{tiny}±0.0\end{tiny}, 36.0\begin{tiny}±0.0\end{tiny}, 29.0\begin{tiny}±0.0\end{tiny})   
& (50.0\begin{tiny}±0.0\end{tiny}, 4.2\begin{tiny}±0.0\end{tiny}, 2.1\begin{tiny}±0.0\end{tiny}) 
& (49.9\begin{tiny}±0.0\end{tiny}, 5.4\begin{tiny}±0.0\end{tiny}, 2.3\begin{tiny}±0.0\end{tiny}) 
& (68.9\begin{tiny}±0.0\end{tiny}, 14.3\begin{tiny}±0.0\end{tiny}, \textbf{8.4}\begin{tiny}±0.0\end{tiny}) 
& (\textbf{74.0}\begin{tiny}±0.9\end{tiny}, \textbf{15.0}\begin{tiny}±0.9\end{tiny}, \underline{7.8}\begin{tiny}±0.8\end{tiny}) 
& (\underline{72.0}\begin{tiny}±0.8\end{tiny}, \underline{14.8}\begin{tiny}±0.9\end{tiny}, 7.5\begin{tiny}±0.6\end{tiny})                                                                     \\
Foam     
& (79.0\begin{tiny}±0.0\end{tiny}, 35.4\begin{tiny}±0.0\end{tiny}, 29.5\begin{tiny}±0.0\end{tiny})   
& (50.0\begin{tiny}±0.0\end{tiny}, 1.2\begin{tiny}±0.0\end{tiny}, 0.6\begin{tiny}±0.0\end{tiny}) 
& (50.2\begin{tiny}±0.0\end{tiny}, 1.3\begin{tiny}±0.0\end{tiny}, 5.0\begin{tiny}±0.0\end{tiny}) 
& (62.5\begin{tiny}±0.0\end{tiny}, 13.1\begin{tiny}±0.0\end{tiny}, 6.0\begin{tiny}±0.0\end{tiny}) 
& (\underline{67.8}\begin{tiny}±1.9\end{tiny}, \underline{20.1}\begin{tiny}±0.3\end{tiny}, \underline{9.7}\begin{tiny}±0.1\end{tiny}) 
& (\textbf{70.2}\begin{tiny}±1.0\end{tiny}, \textbf{23.8}\begin{tiny}±0.1\end{tiny}, \textbf{12.3}\begin{tiny}±0.2\end{tiny})                                                                    \\
Peach      
& (98.7\begin{tiny}±0.0\end{tiny}, 56.4\begin{tiny}±0.0\end{tiny}, 55.9\begin{tiny}±0.0\end{tiny})   
& (50.0\begin{tiny}±0.0\end{tiny}, 2.6\begin{tiny}±0.0\end{tiny}, 1.3\begin{tiny}±0.0\end{tiny}) 
& (77.8\begin{tiny}±0.0\end{tiny}, 11.0\begin{tiny}±0.0\end{tiny}, 5.3\begin{tiny}±0.0\end{tiny})
& (94.4\begin{tiny}±0.0\end{tiny}, \underline{46.4}\begin{tiny}±0.0\end{tiny}, 30.6\begin{tiny}±0.0\end{tiny})
& (\underline{96.8}\begin{tiny}±0.1\end{tiny}, 40.8\begin{tiny}±0.4\end{tiny}, \underline{31.0}\begin{tiny}±1.0\end{tiny}) 
& (\textbf{98.5}\begin{tiny}±0.1\end{tiny}, \textbf{47.0}\begin{tiny}±0.7\end{tiny}, \textbf{40.3}\begin{tiny}±1.2\end{tiny})                                                                    \\
Potato   
& (97.2\begin{tiny}±0.0\end{tiny}, 43.2\begin{tiny}±0.0\end{tiny}, 39.3\begin{tiny}±0.0\end{tiny})     
& (48.4\begin{tiny}±0.0\end{tiny}, 3.6\begin{tiny}±0.0\end{tiny}, 1.8\begin{tiny}±0.0\end{tiny}) 
& (71.1\begin{tiny}±0.0\end{tiny}, 9.2\begin{tiny}±0.0\end{tiny}, 4.4\begin{tiny}±0.0\end{tiny})  
& (96.3\begin{tiny}±0.0\end{tiny}, \textbf{52.1}\begin{tiny}±0.0\end{tiny}, 42.7\begin{tiny}±0.0\end{tiny})
& (\underline{96.8}\begin{tiny}±0.1\end{tiny}, 47.0\begin{tiny}±0.8\end{tiny}, \underline{47.3}\begin{tiny}±2.0\end{tiny})
& (\textbf{98.2}\begin{tiny}±0.0\end{tiny}, \underline{49.0}\begin{tiny}±0.4\end{tiny}, \textbf{49.9}\begin{tiny}±0.3\end{tiny})                                                                    \\
Rope      
& (96.8\begin{tiny}±0.0\end{tiny}, 51.2\begin{tiny}±0.0\end{tiny}, 48.5\begin{tiny}±0.0\end{tiny})    
& (53.9\begin{tiny}±0.0\end{tiny}, 10.6\begin{tiny}±0.0\end{tiny}, 3.8\begin{tiny}±0.0\end{tiny})
& (85.8\begin{tiny}±0.0\end{tiny}, 24.8\begin{tiny}±0.0\end{tiny}, 14.7\begin{tiny}±0.0\end{tiny})
& (68.6\begin{tiny}±0.0\end{tiny}, 14.1\begin{tiny}±0.0\end{tiny}, 8.0\begin{tiny}±0.0\end{tiny})
& (\underline{94.7}\begin{tiny}±0.1\end{tiny}, \underline{47.2}\begin{tiny}±0.5\end{tiny}, \underline{39.7}\begin{tiny}±0.5\end{tiny})
& (\textbf{96.6}\begin{tiny}±0.1\end{tiny}, \textbf{51.5}\begin{tiny}±0.5\end{tiny}, \textbf{44.7}\begin{tiny}±0.4\end{tiny})                                                                    \\
Tire       
& (96.9\begin{tiny}±0.0\end{tiny}, 42.4\begin{tiny}±0.0\end{tiny}, 38.1\begin{tiny}±0.0\end{tiny})     
& (50.0\begin{tiny}±0.0\end{tiny}, 1.6\begin{tiny}±0.0\end{tiny}, 0.8\begin{tiny}±0.0\end{tiny}) 
& (47.8\begin{tiny}±0.0\end{tiny}, 1.8\begin{tiny}±0.0\end{tiny}, \underline{7.0}\begin{tiny}±0.0\end{tiny})  
& (73.3\begin{tiny}±0.0\end{tiny}, 5.3\begin{tiny}±0.0\end{tiny}, 2.2\begin{tiny}±0.0\end{tiny})  
& (\underline{78.5}\begin{tiny}±0.5\end{tiny}, \underline{11.3}\begin{tiny}±0.3\end{tiny}, 4.5\begin{tiny}±0.2\end{tiny}) 
& (\textbf{81.5}\begin{tiny}±0.5\end{tiny}, \textbf{15.7}\begin{tiny}±0.2\end{tiny}, \textbf{7.5}\begin{tiny}±0.2\end{tiny})                                                                   \\ \midrule  
Mean        
& (93.5\begin{tiny}±0.0\end{tiny}, 45.3\begin{tiny}±0.0\end{tiny}, 42.1\begin{tiny}±0.0\end{tiny})    
& (50.3\begin{tiny}±0.0\end{tiny}, 4.0\begin{tiny}±0.0\end{tiny}, 1.9\begin{tiny}±0.0\end{tiny}) 
& (64.7\begin{tiny}±0.0\end{tiny}, 8.6\begin{tiny}±0.0\end{tiny}, 4.7\begin{tiny}±0.0\end{tiny})  
& ({77.6}\begin{tiny}±0.0\end{tiny}, {25.8}\begin{tiny}±0.0\end{tiny}, {18.7}\begin{tiny}±0.0\end{tiny})
& (\underline{85.9}\begin{tiny}±0.3\end{tiny}, \underline{30.5}\begin{tiny}±0.3\end{tiny}, \underline{23.6}\begin{tiny}±0.4\end{tiny}) 
& \begin{tabular}[c]{@{}c@{}}(\textbf{87.5}\begin{tiny}±0.2\end{tiny}, \textbf{33.6}\begin{tiny}±0.1\end{tiny}, \textbf{26.9}\begin{tiny}±0.3\end{tiny})\\ (\textcolor{red}{\textbf{↑1.6, ↑3.1, ↑3.3}})\end{tabular} \\ \bottomrule[1.5pt]

\end{tabular}
}
\label{table:pointmvtec}
\end{table*}

\begin{table*}[]
\setlength{\tabcolsep}{8pt}
\centering
\caption{Quantitative Point-wise Results on Real3D-ad Dataset. The Three Indicators In Parentheses Are P-R, P-F And P-P, Respectively. The Best Is In \textbf{Bold}, And The Second Best In \underline{Underlined}. (\%)}
\setlength\tabcolsep{5.0pt}
\resizebox{\linewidth}{!}{
\begin{tabular}{@{}ccccccc@{}}
\toprule[1.5pt]

Category   & \begin{tabular}[c]{@{}c@{}}CPMF (Unsupervised)~\cite{cpmf}\\ \textit{PR'2024}\end{tabular}
& \begin{tabular}[c]{@{}c@{}}PointMAE~\cite{pointmae}\\ \textit{ECCV'2022}\end{tabular}
& \begin{tabular}[c]{@{}c@{}}MVP-WinCLIP~\cite{winclip}\\ \textit{CVPR'2023}\end{tabular}
& \begin{tabular}[c]{@{}c@{}}MVP-SAA~\cite{saa}\\ \textit{CVPRW'2023}\end{tabular}
& \begin{tabular}[c]{@{}c@{}}MVP-APRIL-GAN~\cite{vand}\\ \textit{CVPRW'2023}\end{tabular}
& \ourmethod{} \\ \midrule

Airplane 
& (61.8\begin{tiny}±0.0\end{tiny}, 2.3\begin{tiny}±0.0\end{tiny}, 1.0\begin{tiny}±0.0\end{tiny})   
& (50.0\begin{tiny}±0.0\end{tiny}, 1.6\begin{tiny}±0.0\end{tiny}, 0.8\begin{tiny}±0.0\end{tiny}) 
& (42.4\begin{tiny}±0.0\end{tiny}, 2.6\begin{tiny}±0.0\end{tiny}, 0.7\begin{tiny}±0.0\end{tiny})  
& (73.7\begin{tiny}±0.0\end{tiny}, \underline{12.6}\begin{tiny}±0.0\end{tiny}, 2.8\begin{tiny}±0.0\end{tiny})  
& (\underline{78.2}\begin{tiny}±0.7\end{tiny}, 9.5\begin{tiny}±2.6\end{tiny}, \underline{3.3}\begin{tiny}±0.6\end{tiny})   
& (\textbf{81.7}\begin{tiny}±0.5\end{tiny}, \textbf{16.4}\begin{tiny}±1.8\end{tiny}, \textbf{10.1}\begin{tiny}±1.6\end{tiny}) \\
Candybar 
& (83.6\begin{tiny}±0.0\end{tiny}, 13.5\begin{tiny}±0.0\end{tiny}, 6.4\begin{tiny}±0.0\end{tiny}) 
& (50.0\begin{tiny}±0.0\end{tiny}, 2.7\begin{tiny}±0.0\end{tiny}, 1.4\begin{tiny}±0.0\end{tiny})
& (57.5\begin{tiny}±0.0\end{tiny}, 5.8\begin{tiny}±0.0\end{tiny}, 1.9\begin{tiny}±0.0\end{tiny}) 
& (70.4\begin{tiny}±0.0\end{tiny}, 6.1\begin{tiny}±0.0\end{tiny}, 2.4\begin{tiny}±0.0\end{tiny})  
& (\underline{76.6}\begin{tiny}±1.7\end{tiny}, \underline{8.6}\begin{tiny}±1.5\end{tiny}, \underline{3.1}\begin{tiny}±0.6\end{tiny}) 
& (\textbf{79.8}\begin{tiny}±0.4\end{tiny}, \textbf{13.9}\begin{tiny}±2.3\end{tiny}, \textbf{6.4}\begin{tiny}±0.8\end{tiny})  \\
Car     
& (73.4\begin{tiny}±0.0\end{tiny}, 10.7\begin{tiny}±0.0\end{tiny}, 5.0\begin{tiny}±0.0\end{tiny}) 
& (50.0\begin{tiny}±0.0\end{tiny}, 2.3\begin{tiny}±0.0\end{tiny}, 1.2\begin{tiny}±0.0\end{tiny}) 
& (50.8\begin{tiny}±0.0\end{tiny}, \underline{7.0}\begin{tiny}±0.0\end{tiny}, \underline{1.8}\begin{tiny}±0.0\end{tiny}) 
& (58.6\begin{tiny}±0.0\end{tiny}, 4.0\begin{tiny}±0.0\end{tiny}, 1.6\begin{tiny}±0.0\end{tiny})  
& (62.8\begin{tiny}±2.4\end{tiny}, 4.3\begin{tiny}±0.7\end{tiny}, 1.4\begin{tiny}±0.3\end{tiny})  
& (\textbf{69.5}\begin{tiny}±1.0\end{tiny}, \textbf{7.4}\begin{tiny}±0.6\end{tiny}, \textbf{4.5}\begin{tiny}±0.7\end{tiny})   \\
Chicken 
& (55.9\begin{tiny}±0.0\end{tiny}, 7.1\begin{tiny}±0.0\end{tiny}, 3.1\begin{tiny}±0.0\end{tiny})  
& (50.0\begin{tiny}±0.0\end{tiny}, 5.5\begin{tiny}±0.0\end{tiny}, 2.8\begin{tiny}±0.0\end{tiny}) 
& (60.3\begin{tiny}±0.0\end{tiny}, 8.6\begin{tiny}±0.0\end{tiny}, 3.9\begin{tiny}±0.0\end{tiny})  
& (\textbf{90.6}\begin{tiny}±0.0\end{tiny}, \textbf{41.6}\begin{tiny}±0.0\end{tiny}, \textbf{32.2}\begin{tiny}±0.0\end{tiny})
& (87.3\begin{tiny}±0.9\end{tiny}, 29.3\begin{tiny}±1.7\end{tiny}, 21.6\begin{tiny}±2.7\end{tiny})
& (\underline{88.8}\begin{tiny}±0.3\end{tiny}, \underline{32.2}\begin{tiny}±0.7\end{tiny}, \underline{27.8}\begin{tiny}±0.9\end{tiny}) \\
Diamond  
& (75.3\begin{tiny}±0.0\end{tiny}, 14.9\begin{tiny}±0.0\end{tiny}, 7.4\begin{tiny}±0.0\end{tiny}) 
& (50.0\begin{tiny}±0.0\end{tiny}, 5.5\begin{tiny}±0.0\end{tiny}, 2.8\begin{tiny}±0.0\end{tiny})
& (68.5\begin{tiny}±0.0\end{tiny}, 9.1\begin{tiny}±0.0\end{tiny}, 4.7\begin{tiny}±0.0\end{tiny})  
& (\underline{90.7}\begin{tiny}±0.0\end{tiny}, 31.3\begin{tiny}±0.0\end{tiny}, 18.8\begin{tiny}±0.0\end{tiny})
& (90.0\begin{tiny}±2.0\end{tiny}, \underline{38.0}\begin{tiny}±1.4\end{tiny}, \underline{33.5}\begin{tiny}±2.2\end{tiny}) 
& (\textbf{95.3}\begin{tiny}±0.9\end{tiny}, \textbf{48.5}\begin{tiny}±2.0\end{tiny}, \textbf{47.9}\begin{tiny}±2.3\end{tiny}) \\
Duck   
& (71.9\begin{tiny}±0.0\end{tiny}, 4.2\begin{tiny}±0.0\end{tiny}, 1.8\begin{tiny}±0.0\end{tiny})  
& (50.0\begin{tiny}±0.0\end{tiny}, 1.9\begin{tiny}±0.0\end{tiny}, 1.0\begin{tiny}±0.0\end{tiny}) 
& (44.3\begin{tiny}±0.0\end{tiny}, 2.0\begin{tiny}±0.0\end{tiny}, 0.8\begin{tiny}±0.0\end{tiny}) 
& (80.5\begin{tiny}±0.0\end{tiny}, \underline{11.1}\begin{tiny}±0.0\end{tiny}, 3.7\begin{tiny}±0.0\end{tiny})
& (\underline{86.6}\begin{tiny}±0.6\end{tiny}, 10.4\begin{tiny}±1.1\end{tiny}, \underline{6.4}\begin{tiny}±1.3\end{tiny}) 
& (\textbf{88.1}\begin{tiny}±0.7\end{tiny}, \textbf{14.0}\begin{tiny}±1.4\end{tiny}, \textbf{8.2}\begin{tiny}±0.6\end{tiny})  \\
Fish  
& (98.8\begin{tiny}±0.0\end{tiny}, 58.2\begin{tiny}±0.0\end{tiny}, 55.9\begin{tiny}±0.0\end{tiny})
& (50.0\begin{tiny}±0.0\end{tiny}, 2.7\begin{tiny}±0.0\end{tiny}, 1.4\begin{tiny}±0.0\end{tiny}) 
& (71.4\begin{tiny}±0.0\end{tiny}, 15.4\begin{tiny}±0.0\end{tiny}, 6.3\begin{tiny}±0.0\end{tiny})
& (\textbf{93.1}\begin{tiny}±0.0\end{tiny}, \textbf{67.9}\begin{tiny}±0.0\end{tiny}, \textbf{68.5}\begin{tiny}±0.0\end{tiny})
& (82.4\begin{tiny}±2.4\end{tiny}, \underline{28.6}\begin{tiny}±0.7\end{tiny}, \underline{17.4}\begin{tiny}±0.2\end{tiny}) 
& (\underline{85.8}\begin{tiny}±1.6\end{tiny}, 18.6\begin{tiny}±1.7\end{tiny}, 11.0\begin{tiny}±1.1\end{tiny}) \\
Gemstone 
& (44.9\begin{tiny}±0.0\end{tiny}, 2.0\begin{tiny}±0.0\end{tiny}, 0.7\begin{tiny}±0.0\end{tiny}) 
& (50.0\begin{tiny}±0.0\end{tiny}, 1.8\begin{tiny}±0.0\end{tiny}, 1.8\begin{tiny}±0.0\end{tiny})
& (66.5\begin{tiny}±0.0\end{tiny}, 3.1\begin{tiny}±0.0\end{tiny}, 1.3\begin{tiny}±0.0\end{tiny}) 
& (79.5\begin{tiny}±0.0\end{tiny}, 7.9\begin{tiny}±0.0\end{tiny}, 2.8\begin{tiny}±0.0\end{tiny})  
& (\underline{87.9}\begin{tiny}±1.0\end{tiny}, \underline{24.0}\begin{tiny}±2.0\end{tiny}, \underline{13.3}\begin{tiny}±1.0\end{tiny})
& (\textbf{91.5}\begin{tiny}±0.4\end{tiny}, \textbf{28.0}\begin{tiny}±0.5\end{tiny}, \textbf{18.2}\begin{tiny}±0.8\end{tiny}) \\
Seahorse
& (96.2\begin{tiny}±0.0\end{tiny}, 61.5\begin{tiny}±0.0\end{tiny}, 63.6\begin{tiny}±0.0\end{tiny})
& (50.0\begin{tiny}±0.0\end{tiny}, 4.9\begin{tiny}±0.0\end{tiny}, 4.8\begin{tiny}±0.0\end{tiny})
& (65.3\begin{tiny}±0.0\end{tiny}, 8.7\begin{tiny}±0.0\end{tiny}, 4.1\begin{tiny}±0.0\end{tiny}) 
& (64.3\begin{tiny}±0.0\end{tiny}, 16.1\begin{tiny}±0.0\end{tiny}, 11.1\begin{tiny}±0.0\end{tiny}) 
& (\textbf{81.5}\begin{tiny}±1.9\end{tiny}, \textbf{24.8}\begin{tiny}±4.2\end{tiny}, \textbf{18.3}\begin{tiny}±3.1\end{tiny}) 
& (\underline{80.5}\begin{tiny}±1.2\end{tiny}, \underline{21.7}\begin{tiny}±3.4\end{tiny}, \underline{17.1}\begin{tiny}±3.1\end{tiny}) \\
Shell   
& (72.5\begin{tiny}±0.0\end{tiny}, 5.2\begin{tiny}±0.0\end{tiny}, 2.5\begin{tiny}±0.0\end{tiny}) 
& (50.0\begin{tiny}±0.0\end{tiny}, 2.3\begin{tiny}±0.0\end{tiny}, 2.3\begin{tiny}±0.0\end{tiny})
& (62.0\begin{tiny}±0.0\end{tiny}, 9.4\begin{tiny}±0.0\end{tiny}, 2.7\begin{tiny}±0.0\end{tiny}) 
& (87.3\begin{tiny}±0.0\end{tiny}, 19.1\begin{tiny}±0.0\end{tiny}, \textbf{17.0}\begin{tiny}±0.0\end{tiny})
& (\underline{89.1}\begin{tiny}±1.4\end{tiny}, \underline{26.3}\begin{tiny}±1.6\end{tiny}, 12.5\begin{tiny}±1.0\end{tiny}) 
& (\textbf{90.1}\begin{tiny}±0.7\end{tiny}, \textbf{28.1}\begin{tiny}±1.4\end{tiny}, \underline{16.4}\begin{tiny}±1.0\end{tiny}) \\
Starfish 
& (80.0\begin{tiny}±0.0\end{tiny}, 20.2\begin{tiny}±0.0\end{tiny}, 12.8\begin{tiny}±0.0\end{tiny}) 
& (50.0\begin{tiny}±0.0\end{tiny}, 4.5\begin{tiny}±0.0\end{tiny}, 4.5\begin{tiny}±0.0\end{tiny})
& (63.1\begin{tiny}±0.0\end{tiny}, 7.1\begin{tiny}±0.0\end{tiny}, 3.7\begin{tiny}±0.0\end{tiny})  
& (59.7\begin{tiny}±0.0\end{tiny}, 14.7\begin{tiny}±0.0\end{tiny}, 6.0\begin{tiny}±0.0\end{tiny}) 
& (\textbf{75.4}\begin{tiny}±0.6\end{tiny}, \textbf{22.2}\begin{tiny}±2.6\end{tiny}, \textbf{13.4}\begin{tiny}±2.0\end{tiny})
& (\underline{70.9}\begin{tiny}±1.2\end{tiny}, \underline{19.4}\begin{tiny}±0.8\end{tiny}, \underline{12.1}\begin{tiny}±0.5\end{tiny}) \\
Toffees
& (95.9\begin{tiny}±0.0\end{tiny}, 46.0\begin{tiny}±0.0\end{tiny}, 39.1\begin{tiny}±0.0\end{tiny})
& (50.0\begin{tiny}±0.0\end{tiny}, 2.2\begin{tiny}±0.0\end{tiny}, 2.2\begin{tiny}±0.0\end{tiny}) 
& (48.1\begin{tiny}±0.0\end{tiny}, 3.0\begin{tiny}±0.0\end{tiny}, 1.1\begin{tiny}±0.0\end{tiny}) 
& (87.4\begin{tiny}±0.0\end{tiny}, 18.8\begin{tiny}±0.0\end{tiny}, 8.6\begin{tiny}±0.0\end{tiny}) 
& (\textbf{94.8}\begin{tiny}±0.3\end{tiny}, \textbf{41.1}\begin{tiny}±5.6\end{tiny}, \textbf{37.5}\begin{tiny}±3.7\end{tiny})
& (\underline{93.4}\begin{tiny}±0.5\end{tiny}, \underline{38.7}\begin{tiny}±3.7\end{tiny}, \underline{34.7}\begin{tiny}±2.5\end{tiny})                                                               \\ \midrule 
Mean     
& (75.8\begin{tiny}±0.0\end{tiny}, 20.5\begin{tiny}±0.0\end{tiny}, 16.6\begin{tiny}±0.0\end{tiny})
& (50.0\begin{tiny}±0.0\end{tiny}, 3.2\begin{tiny}±0.0\end{tiny}, 2.3\begin{tiny}±0.0\end{tiny}) 
& (58.4\begin{tiny}±0.0\end{tiny}, 6.8\begin{tiny}±0.0\end{tiny}, 2.7\begin{tiny}±0.0\end{tiny})  
& (78.0\begin{tiny}±0.0\end{tiny}, 20.9\begin{tiny}±0.0\end{tiny}, 14.6\begin{tiny}±0.0\end{tiny}) 
& (\underline{82.7}\begin{tiny}±1.0\end{tiny}, \underline{22.3}\begin{tiny}±1.2\end{tiny}, \underline{15.1}\begin{tiny}±1.0\end{tiny}) 
&  \begin{tabular}[c]{@{}c@{}}(\textbf{84.6}\begin{tiny}±0.3\end{tiny}, \textbf{23.9}\begin{tiny}±0.4\end{tiny}, \textbf{17.9}\begin{tiny}±0.1\end{tiny})\\ (\textcolor{red}{\textbf{↑1.9, ↑1.6, ↑2.8}})\end{tabular} \\

\bottomrule[1.5pt]

\end{tabular}
}
\label{table:pointreal}
\end{table*}

\begin{figure*}[t!]
\centering\includegraphics[width=\linewidth]{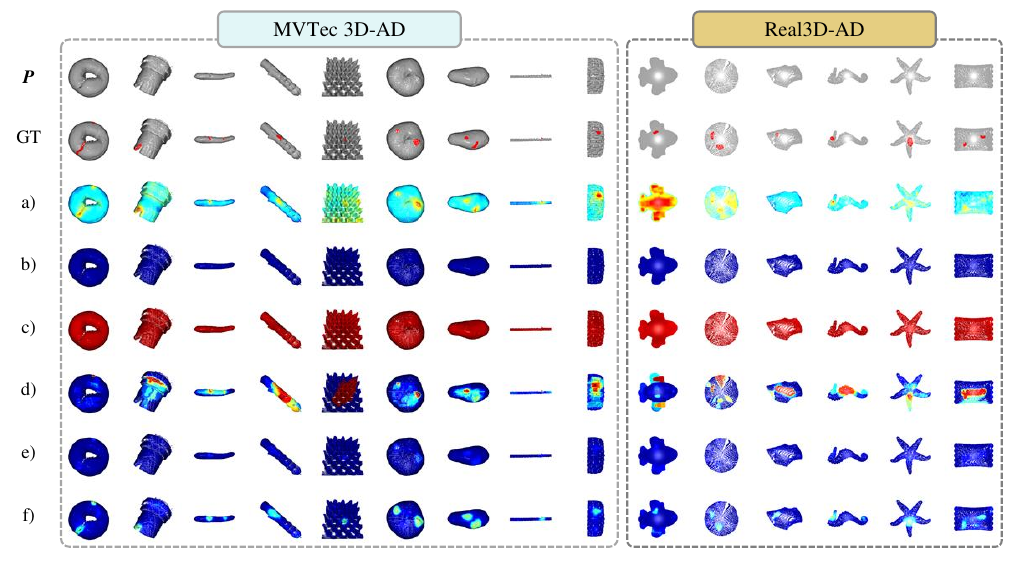}
\caption{Visualization of prediction results using the proposed method and other methods. The first nine columns are from MVTec 3D-AD dataset, and the last six columns are from Real3D-AD dataset. The first row is the original point clouds, while the second row is the ground truth. Subsequent rows depict various methods: a) CPMF b) PointMAE c) MVP-WinCLIP d) MVP-SAA e) MVP-APRIL-GAN f) MVP-PCLIP.}
\label{fig:Visualization}
\end{figure*}

As shown in Fig.~\ref{fig:Visualization}, our method effectively identifies anomalies across various categories in both datasets, showcasing a notable reduction in missed detections and false positives. While CPMF excels in point-wise performance on the MVTec 3D-AD dataset, it encounters a significant drawback on the Real3D-AD dataset, generating numerous false positives attributed to misaligned training sample features. PointMAE and MVP-WinCLIP fail in zero-shot point cloud anomaly detection task. PointMAE incorrectly treats all points as normal, while MVP-WinCLIP categorizes all points as anomalies. Despite MVP-SAA's robust segmentation ability, it struggles to comprehend the semantics of each part of the point cloud, resulting in a considerable number of false detections. In contrast, MVP-APRIL-GAN can effectively detect most defects but with relatively low anomaly scores. It is worth noting that MVP-PCLIP not only outperforms other zero-shot anomaly detection methods on the MVTec 3D-AD dataset but also surpasses all other methods when applied to the more intricate point clouds in the Real3D-AD dataset, benefiting by bridging the gap between training and target datasets.

\subsubsection{Object-wise Anomaly Detection}
Quantitative results for the MVTec 3D-AD dataset and Real3D-AD dataset are presented in Table~\ref{table:object}. On the MVTec 3D-AD dataset, MVP-PCLIP exhibits the best performance in zero-shot anomaly detection, achieving +0.9\% higher O-R, +1.6\% higher O-F, and +0.4\% higher O-P compared to the second-placed WinCLIP. In the Real3D-AD dataset, MVP-PCLIP again leads in zero-shot anomaly detection, performing +1.2\% higher O-R, +0.3\% higher O-F, and +2.8\% higher O-P than the second preference. Notably, our method even surpasses the unsupervised method CPMF in O-F.

Visualizations of point-wise results in Fig.~\ref{fig:Visualization} reveal that all methods, except PointMAE, tend to classify all objects as normal. Table~\ref{table:object} further indicates that the O-R of MVP-SAA and MVP-APRIL-GAN hovers around 50\%, indicating an inability to distinguish between normal and abnormal objects. Unsupervised methods heavily rely on the abundance of normal samples, and as a result, CPMF performs admirably on the MVTec 3D-AD dataset with ample training data. Conversely, the limited training data in the Real3D-AD dataset significantly diminishes CPMF's performance. Overall, zero-shot anomaly detection methods exhibit poor performance due to pronounced over-detection.

The inclination towards over-detection arises from the greater dissimilarity between two datasets than the dissimilarity between normal and abnormal data within the dataset, which exacerbates the model's tendency to classify all test data as abnormal. MVP-PCLIP, while displaying a similar inclination, mitigates this issue through the incorporation of visual prompts to address domain gaps and adaptive text prompts for improved text descriptions, resulting in superior performance.

\begin{table}[]
\setlength{\tabcolsep}{8pt}
\centering
\caption{Quantitative Object-wise Results On The MVTec 3D-AD Dataset and Real3D-AD Dataset. The Three Indicators In Parentheses Are O-R, O-F And O-P, Respectively. The Best Is In \textbf{Bold}, And The Second Best In \underline{Underlined}.(\%)}
\setlength\tabcolsep{5.0pt}
\resizebox{\linewidth}{!}{
\begin{tabular}{@{}ccc@{}}
\toprule[1.5pt]

Category   & MVTec 3D-AD & Real3D-AD \\  \midrule
\begin{tabular}[c]{@{}c@{}}CPMF (Unsupervised)~\cite{cpmf}\\ \textit{PR'2024}\end{tabular} & (94.6\begin{tiny}±0.0\end{tiny}, 98.6\begin{tiny}±0.0\end{tiny}, 95.2\begin{tiny}±0.0\end{tiny}) & (62.5\begin{tiny}±0.0\end{tiny}, 64.2\begin{tiny}±0.0\end{tiny}, 72.3\begin{tiny}±0.0\end{tiny}) \\

\begin{tabular}[c]{@{}c@{}}PointMAE~\cite{pointmae}\\ \textit{ECCV'2022}\end{tabular} & (4.7\begin{tiny}±0.0\end{tiny}, 88.2\begin{tiny}±0.0\end{tiny}, 78.5\begin{tiny}±0.0\end{tiny}) & (51.4\begin{tiny}±0.0\end{tiny}, 67.8\begin{tiny}±0.0\end{tiny}, \underline{54.7}\begin{tiny}±0.0\end{tiny})\\

\begin{tabular}[c]{@{}c@{}}MVP-WinCLIP~\cite{winclip}\\ \textit{CVPR'2023}\end{tabular} &  (\underline{71.0}\begin{tiny}±0.0\end{tiny}, \underline{89.1}\begin{tiny}±0.0\end{tiny}, \underline{89.2}\begin{tiny}±0.0\end{tiny})   & (\underline{54.1}\begin{tiny}±0.0\end{tiny}, 68.5\begin{tiny}±0.0\end{tiny}, 54.5\begin{tiny}±0.0\end{tiny}) \\

\begin{tabular}[c]{@{}c@{}}MVP-SAA~\cite{saa}\\ \textit{CVPRW'2023}\end{tabular} & (54.7\begin{tiny}±0.0\end{tiny}, 88.4\begin{tiny}±0.0\end{tiny}, 81.8\begin{tiny}±0.0\end{tiny}) & (49.1\begin{tiny}±0.0\end{tiny}, \underline{69.2}\begin{tiny}±0.0\end{tiny}, 54.5\begin{tiny}±0.0\end{tiny}) \\

\begin{tabular}[c]{@{}c@{}}MVP-APRIL-GAN~\cite{vand}\\ \textit{CVPRW'2023}\end{tabular} & (66.9\begin{tiny}±8.5\end{tiny}, 88.2\begin{tiny}±0.5\end{tiny}, 82.6\begin{tiny}±4.2\end{tiny}) &  (51.2\begin{tiny}±4.0\end{tiny}, 68.1\begin{tiny}±1.0\end{tiny}, 54.2\begin{tiny}±2.8\end{tiny})\\ \midrule

\ourmethod{}  &  \begin{tabular}[c]{@{}c@{}}(\textbf{71.9}\begin{tiny}±2.4\end{tiny}, \textbf{90.7}\begin{tiny}±0.1\end{tiny}, \textbf{89.6}\begin{tiny}±1.8\end{tiny})\\ (\textcolor{red}{\textbf{↑0.9, ↑1.6, ↑0.4}})\end{tabular}  & \begin{tabular}[c]{@{}c@{}}(\textbf{55.3}\begin{tiny}±0.8\end{tiny}, \textbf{69.5}\begin{tiny}±0.2\end{tiny}, \textbf{57.5}\begin{tiny}±1.4\end{tiny})\\ (\textcolor{red}{\textbf{↑1.2, ↑0.3, ↑2.8}})\end{tabular}  

\\

\bottomrule[1.5pt]

\end{tabular}
}
\label{table:object}
\end{table}

\subsection{Ablation Studies}

In this subsection, \cyq{we investigate the effects of the proposed visual and text prompts, as well as the optimal parameter settings}. Additionally, we examine the influence of the number of views on both performance and processing speed. Finally, we analyze the impact of the backbone architecture.

\subsubsection{Influence of prompt}

The point-wise detection results, obtained through various prompt combinations, are presented in Table~\ref{table:submodel}. The term "Base" refers to the detection framework without any prompts. The results demonstrate that both ATP and KLVP contribute to an improvement in performance, with the combination of both prompts achieving the optimal results.

\begin{table}[h!]
\caption{Quantitative Point-wise Results Under Different prompt. The Three Indicators In Parentheses Are P-R, P-F And P-P Respectively. (\%)}
\centering\includegraphics[width=\linewidth]{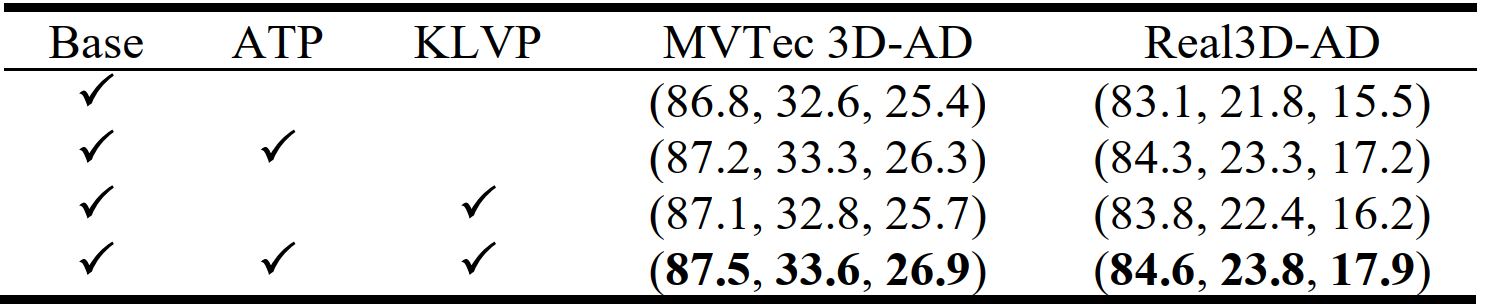}
\label{table:submodel}
\end{table}

\begin{figure}[h!]
\centering\includegraphics[scale=1.0]{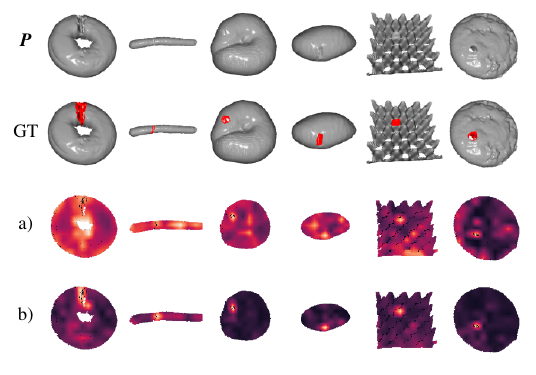}
\caption{The visualizations of extracted features from image encoder. The first row is the original point clouds, while the second row is the ground truth. (a) Extracted features obtained from the image encoder of CLIP. (b) Extracted features obtained from image encoder prompted by KLVP.}
\label{fig:attentionmap}
\end{figure}

Attention maps from key layers of \cyq{the image encoder} before and after incorporating KLVP are showcased in Fig.~\ref{fig:attentionmap}. Those attention maps illustrate the interest that the image encoder assigns to each point. Despite being trained on the auxiliary dataset, as shown in Fig.~\ref{fig:attentionmap} a), the image encoder of CLIP tends to pay fair attention to image regions, resulting in a dispersed attention pattern. Upon integrating KLVP, the image encoder shifts its attention more towards anomalies, thereby enhancing its capability to distinguish abnormal regions, as shown in Fig.~\ref{fig:attentionmap} b).

\subsubsection{Influence of learnable parameter number}
\begin{figure}[h!]
\centering\includegraphics[scale=1.0]{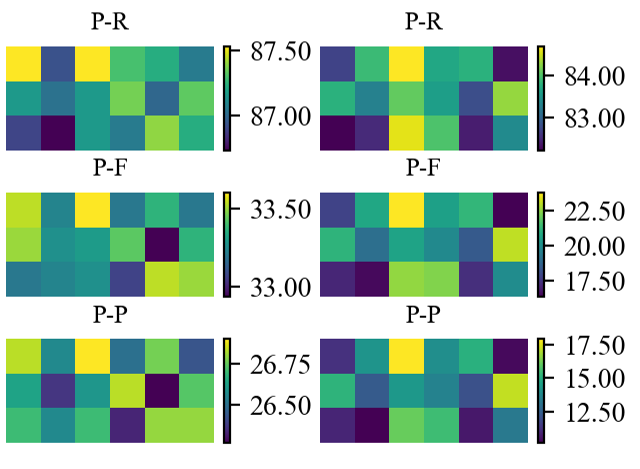}
\caption{The optimization experiment results on MVTec 3D-AD (left) and Real3D-AD (right). The vertical direction of the heatmap represents the number of tokens used by KLVP, and the number of tokens from top to bottom is 1, 2, 4. The horizontal direction represents different settings of union/specific parameters in ATP, and from left to right is 4/2, 4/4, 8/4, 8/8, 16/8, 16/16.}
\label{fig:para}
\end{figure}

Subsequent experiments explore the impact \cyq{on} the number of visual prompt tokens and the union/specific learnable parameters in text prompts. Specifically, the token number of KLVP varies among 1, 2, and 4. Meanwhile, the number of learnable parameters for union/specific text prompts in ATP is explored with configurations of 4/2, 4/4, 8/4, 8/8, 16/8, and 16/16. Multiple experiments are conducted for each setting, and the average performance is illustrated in Fig.~\ref{fig:para}. Integrating KLVP and ATP with different configurations demonstrates performance improvements across all three indicators. Notably, P-F and P-P exhibit more substantial improvements compared to P-R. When the token count for KLVP is fixed, an overall increasing trend followed by a decrease is observed as the number of ATP parameters increases. This pattern emerges due to an insufficient or excessive number of ATP parameters that fail to adequately align with the image features. Conversely, with increasing KLVP parameters, a gradual decline in performance is noted. This occurs because learnable parameters in KLVP alleviate the domain gap but with increasing KLVP parameters, overfitting on auxiliary datasets creates a new domain gap. \cyq{Overall, as depicted in Fig.~\ref{fig:para}, MVP-PCLIP achieves optimal results with a token number of 1 and a union/specific learnable parameter ratio of 8/4.}

\subsubsection{Influence of number of rendering views}
\begin{figure}[h!]
\centering\includegraphics[width=\linewidth]{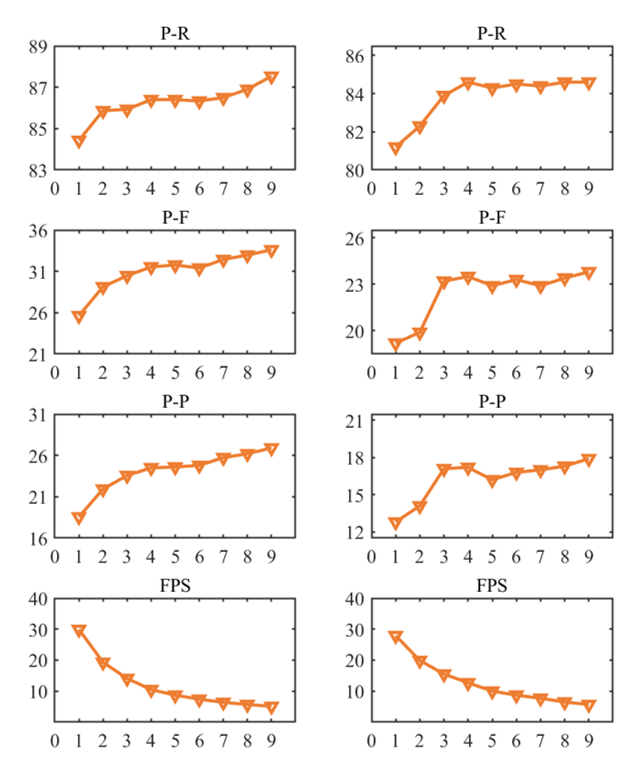}
\caption{The point-wise detection performance and speed using different number of views. The horizontal axis represents the number of different views. (Left) MVTec 3D-AD dataset. (Right) Real3D-AD dataset}
\label{fig:views}
\end{figure}

The saliency of different defect types varies in depth images projected from different angles, so the number of projection views has a significant impact on the performance. \cyq{In Fig.~\ref{fig:views}, the variations in three point-wise metrics and Frames Per Second (FPS) are illustrated as the number of views gradually increases from one to nine. The observations reveal an upward trend in P-R, P-F, and P-P with the increasing number of views, reaching optimal performance when the number of views is set to nine. \cyq{However}, this does not necessarily imply saturation in performance growth according to the variations of metrics. When the number of projected depth images is small, the defect area may be missing or obscured due to the projection angles or projection quality, which leads to poor detection performance. Obtaining a large number of depth images ensures that at least one image contains the required defects, which greatly improves detection performance. As the number of views increases, there is a gradual decrease in FPS, dropping from 29.9 fps and 27.9 fps to 5.1 fps and 5.9 fps. While more views contribute to higher accuracy, the practical implications of reduced efficiency must be carefully considered for real-world applications.}

\subsubsection{Influence of backbones}
To investigate the influence of different backbones on performance, we select ViT-L and ViT-B as the chosen backbones. The evaluation is conducted using point-wise indicators, as depicted in Table.~\ref{table:backbone}. The results showcase a substantial performance difference between ViT-L and ViT-B, with ViT-L outperforming ViT-B. This observation suggests that larger networks, such as ViT-L, excel in extracting more generalized knowledge and better comprehending the nuances associated with anomalies.

\begin{table}[t!]
\caption{The point-wise performance using different backbones. The Three Indicators In Parentheses Are P-R, P-F And P-P Respectively. (\%)}
\centering\includegraphics[width=\linewidth]{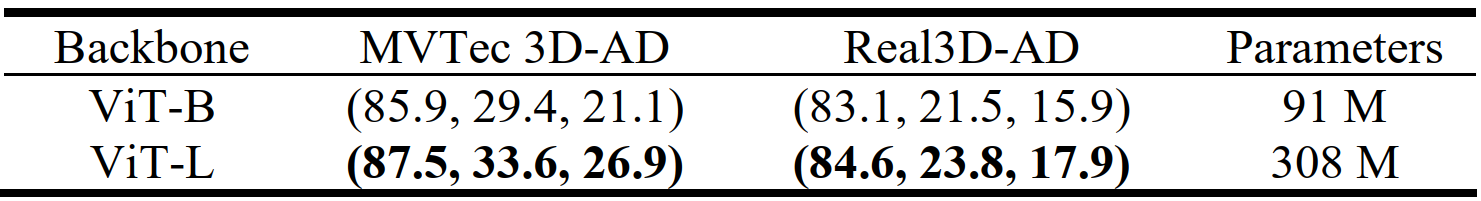}
\label{table:backbone}
\end{table}

\subsection{Failure Case Analysis}

\begin{figure}[t!]
\centering\includegraphics[scale=1.0]{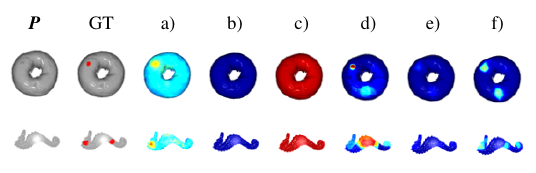}
\caption{Visualization of prediction results of the failure case on MVTec 3D-AD and Real3D-AD. The first row is the original point clouds, while the second row is the ground truth. Subsequent rows depict various methods: a) CPMF b) PointMAE c) MVP-WinCLIP d) MVP-SAA e) MVP-APRIL-GAN f) MVP-PCLIP.}
\label{fig:badcase}
\end{figure}

\cyq{While our method demonstrates commendable performance on public datasets, it exhibits limitations in handling specific samples, as illustrated by the failure cases in Fig.~\ref{fig:badcase}. Challenges emerge when the difference between anomalies and normal areas is subtle, causing the method to erroneously recognize.} Due to the superiority of the MVP framework, the detection results of MVP-SAA, MVP-APRIL-GAN and MVP-PCLIP still cover anomalies, as observed in the first rows of Fig.~\ref{fig:badcase}. Upon examination of the second rows of Fig.~\ref{fig:badcase}, it is evident that while experiencing certain limitations, MVP-PCLIP has a superior ability to identify abnormalities and surpasses other approaches. This is attributed to the incorporation of visual and text prompts.

\begin{figure}[t!]
\centering\includegraphics[scale=1.0]{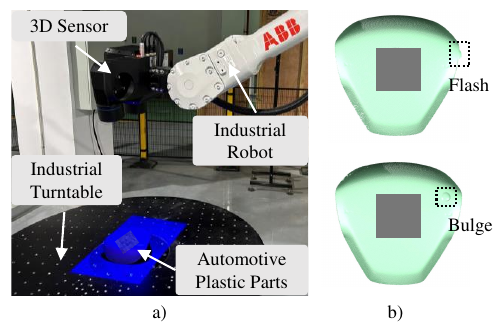}
\caption{a) The point cloud collection process of automotive plastic parts. b) Some abnormal samples with flash or bulge.
}
\label{fig:scan}
\end{figure}

\subsection{Practical Inspection on Automotive Plastic Parts}

\cyq{To further evaluate the actual performance of the proposed MVP-PCLIP, inspection experiments on actual automotive plastic parts are conducted. Specifically, a point cloud acquisition equipment is shown in Fig.~\ref{fig:scan} a), which utilizes an industrial robot equipped with 3D sensors to scan the automotive parts. These parts may comprise numerous anomalies, like flashes and bulges, as shown in Fig.~\ref{fig:scan} b), where the brand logos are masked for confidentiality. The collected dataset includes 50 normal samples and 50 abnormal samples in total, each including about 100000 points. All zero-shot methods mentioned above are evaluated on this dataset, for the methods requiring training, i.e., MVP-PCLIP and MVP-APRIL-GAN, MVTec 3D-AD and Real3D-AD are utilized as auxiliary datasets.}

\begin{table}[]
\setlength{\tabcolsep}{8pt}
\centering
\caption{Quantitative Results on Plastic Parts Inspection. The Three Indicators In Parentheses Are AUROC, F1-Max and AP Respectively. (\%)}
\setlength\tabcolsep{5.0pt}
\resizebox{\linewidth}{!}{
\begin{tabular}{@{}ccc@{}}
\toprule[1.5pt]

Category   & Object-wise & Point-wise \\  \midrule

\begin{tabular}[c]{@{}c@{}}PointMAE~\cite{pointmae}\\ \textit{ECCV'2022}\end{tabular}  & (54.4, 66.7, 59.6) & (50, 0.7, 0.3) \\
\begin{tabular}[c]{@{}c@{}}MVP-WinCLIP~\cite{winclip}\\ \textit{CVPR'2023}\end{tabular} & (55.3, 61.6, 58.3)   & (50, 1.2, 0.3) \\
\begin{tabular}[c]{@{}c@{}}MVP-SAA~\cite{saa}\\ \textit{CVPRW'2023}\end{tabular} & (34.5, 66.7, 45.6) & (83.5, 7.6, 3.1) \\
\begin{tabular}[c]{@{}c@{}}MVP-APRIL-GAN~\cite{vand}\\ \textit{CVPRW'2023}\end{tabular} & (50.9, 67.1, 57.2) &  (88.3, 5.4, 2.6)  \\ \midrule
\ourmethod{}  &  \begin{tabular}[c]{@{}c@{}} (\textbf{57.4}, \textbf{68.5}, \textbf{62.6} \\ (\textcolor{red}{\textbf{↑2.1, ↑1.4, ↑3.0}})\end{tabular}  
& \begin{tabular}[c]{@{}c@{}}(\textbf{93.1}, \textbf{27.9}, \textbf{20.2})\\ (\textcolor{red}{\textbf{↑9.6, ↑20.3, ↑17.1}})\end{tabular}  

\\

\bottomrule[1.5pt]

\end{tabular}
}
\label{table:PP}
\end{table}

\begin{figure}[t!]
\centering\includegraphics[width=\linewidth]{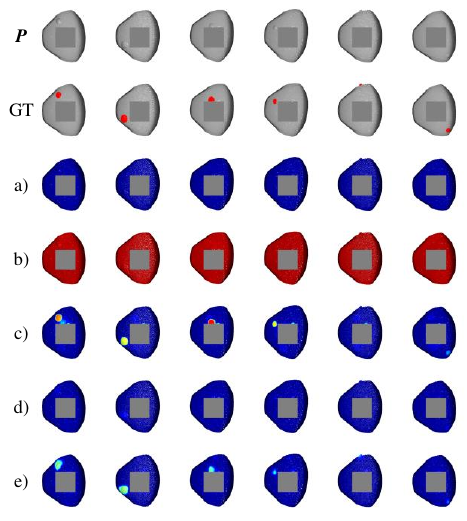}
\caption{Visualization of prediction results using the proposed method and other methods on plastic parts dataset. The first row is the original point clouds, while the second row is the ground truth. Subsequent rows depict various methods: a) PointMAE b) MVP-WinCLIP c) MVP-SAA d) MVP-APRIL-GAN e) MVP-PCLIP.
}
\label{fig:PP}
\end{figure}

\cyq{Table~\ref{table:PP} presents the qualitative comparison results on the collected dataset. It clearly shows that MVP-PCLIP is the best performer in all indicators and significantly outperforms other alternatives, especially in point-wise performance. The visualization results are shown in Fig.~\ref{fig:PP}, which demonstrates that MVP-PCLIP can produce more accurate localization results in comparison to other methods. Although SAA can also detect anomalies well, it mistakenly identifies some normal regions as anomalies. In summary, the proposed MVP-PCLIP achieves excellent performance in zero-shot point cloud anomaly detection on automotive plastic parts and has the potential for practical application.}

\section{Conclusion}\label{sec:conclusion}

This paper introduces a novel and challenging task: zero-shot point cloud anomaly detection. To address this task, we propose a straightforward yet effective framework, Multi-View Projection (MVP). MVP leverages the similarity between point clouds and multi-view images by projecting point clouds into multi-view depth images, thereby extending existing zero-shot image anomaly detection techniques to point clouds. 
To mitigate the domain gap faced by VLMs, we further propose MVP-PCLIP, which incorporates two types of prompts: key layer visual prompt (KLVP) and adaptive text prompt (ATP). KLVP adds learnable parameters to the image encoder, enhancing its focus on anomalous regions, while ATP introduces union/specific learnable parameters to the text description, refining the matching between text and image features.
Extensive experiments demonstrate the effectiveness of the MVP framework, with MVP-PCLIP outperforming previous methods on both the MVTec 3D-AD and Real3D-AD datasets. Ablation studies and practical experiments further validate the feasibility of the proposed method.

\cyq{The proposed MVP-PCLIP is not without its limitations. For example, it is difficult to handle normal and abnormal regions that exhibit subtle differences, which requires stronger associations between text descriptions and point cloud features. Future research on VLMs specifically designed for point clouds will likely improve the performance of zero-shot point cloud anomaly detection. Additionally, generating a sufficient number of abnormal samples could further enhance the generalization capabilities of supervised learning frameworks.}

\section*{Acknowledgments}
The computation is completed in the HPC Platform of Huazhong University of Science and Technology.

\ifCLASSOPTIONcaptionsoff
  \newpage
\fi



\end{document}